\documentclass[final]{cvpr}

\usepackage{times}
\usepackage{epsfig}
\usepackage{graphicx}
\usepackage{amsmath}
\usepackage{amssymb}
\DeclareMathOperator*{\argmax}{arg\,max}
\DeclareMathOperator*{\argmin}{arg\,min}

\usepackage{bbold}
\usepackage{subcaption}
\usepackage{multirow}
\usepackage[font=small,skip=1pt]{caption}

\usepackage{cuted}
\usepackage{capt-of}
\usepackage{array}% http://ctan.org/pkg/array
\makeatletter
\g@addto@macro{\endtabular}{\rowfont{}}% Clear row font
\makeatother
\newcommand{\rowfonttype}{}% Current row font
\newcommand{\rowfont}[1]{% Set current row font
  \gdef\rowfonttype{#1}#1%
}

\usepackage[pagebackref=true,breaklinks=true,colorlinks,bookmarks=false]{hyperref}

\def\cvprPaperID{1299} % *** Enter the CVPR Paper ID here
\def\confYear{CVPR 2021}

\begin{document}

%%%%%%%%% TITLE
\title{IMAGINE: Image Synthesis by Image-Guided Model Inversion}

\author{Pei Wang\thanks{Work done during internship at Adobe Research}\\
UC, San Diego\\
{\tt\small pew062@ucsd.edu}
\and
Yijun Li, Krishna Kumar Singh, Jingwan Lu\\
Adobe Research\\
{\tt\small \{yijli, krishsin, jlu\}@adobe.com}

\and
Nuno Vasconcelos\\
UC, San Diego\\
{\tt\small nuno@ucsd.edu}
}
\maketitle

% \begin{strip}\centering
% \includegraphics[width=1.0\linewidth,height=0.35\textheight]{Inversion/figures/teaser2.pdf}
% \captionof{figure}{Feature graphic caption.}
% \label{fig:teaser2}
% \end{strip}

\begin{strip}\centering
\begin{tabular}{c|c}
\rowfont{\small}
Reference image&\rowfont{\small}Random samples generated by a reference image\\
\includegraphics[width=1.9cm,height=8cm]{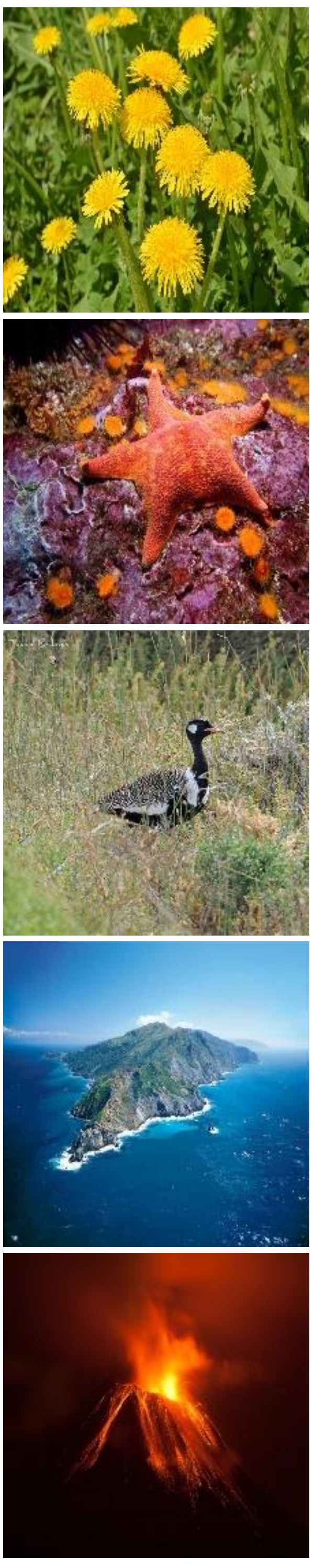}&
\includegraphics[width=14.2cm,height=8.0cm]{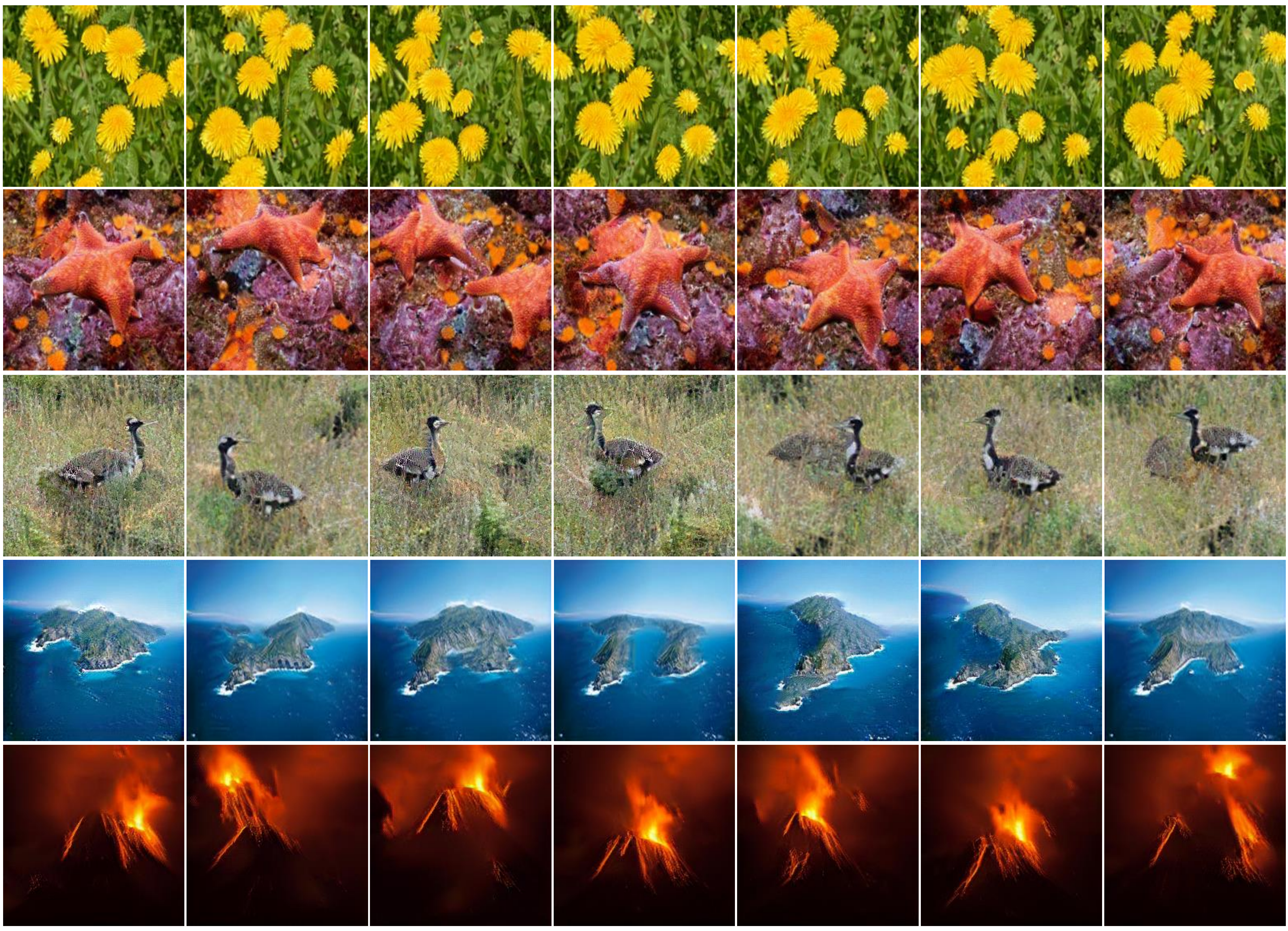}
\end{tabular}
\captionof{figure}{Results from our proposed IMAGINE - the first attempt to use classifier inversion for reference-guided image generation.}
\label{fig:teaser1}
\end{strip}

% scene and object-centerd images, 
% repetative

%%%%%%%%% ABSTRACT
\begin{abstract}

We introduce an inversion based method, denoted as IMAge-Guided model INvErsion (IMAGINE), to generate high-quality and diverse images from only a single training sample. We leverage the knowledge of image semantics from a pre-trained classifier to achieve plausible generations via matching multi-level feature representations in the classifier, associated with adversarial training with an external discriminator.
IMAGINE enables the synthesis procedure to simultaneously 1) enforce semantic specificity constraints during the synthesis, 2) produce realistic images without generator training, and 3) give users intuitive control over the generation process. With extensive experimental results, we demonstrate qualitatively and quantitatively that IMAGINE performs favorably against state-of-the-art GAN-based and inversion-based methods, across three different image domains (i.e., objects, scenes, and textures). 
%IMAGINE is also applied into other various image synthesis related tasks including position control, semantic control and counterfactual explanations. 

\end{abstract}

%%%%%%%%% BODY TEXT
\section{Introduction}

% \cl{should define "fractal" and "non-fractal" in the intro. I tried to do a little bit in Figure 2 caption.}

% \begin{figure*}[t]
% \begin{center}
% \includegraphics[width=1.0\linewidth]{Inversion/figures/teaser2.pdf}
% \end{center}
%   \caption{}
% \label{fig:teaser2}
% \end{figure*}

% \kr{Consider a reference image from the left-most column of Fig.~\ref{fig:teaser1}} 
Consider a reference image from the left-most column of Fig.~\ref{fig:teaser1}, we humans could easily \emph{imagine} its variants in other formats, e.g., patterns with different spatial arrangements, objects in different positions or viewpoints.
Can we give machines such abilities to automatically synthesize semantically meaningful variations of a reference image?
Such a system can generate interesting variations of assets for design ideations and can also be utilized as a data augmentation technique to benefit downstream data-hungry tasks.
%This will be of great importance in terms of personalized generation, which could provide more choices for users based on their personal creation and speed up their editing and design flows.
%
%Moreover, it could also be regarded as a type of effective data augmentation technique to expand the limited amount of data we have at hand and benefit many down-stream data-hungry tasks. 

\begin{figure}[t]
\setlength{\abovecaptionskip}{-2.0pt}
\setlength{\belowcaptionskip}{-0.5pt}
\setlength{\tabcolsep}{2pt}
\begin{center}
\includegraphics[width=1.0\linewidth]{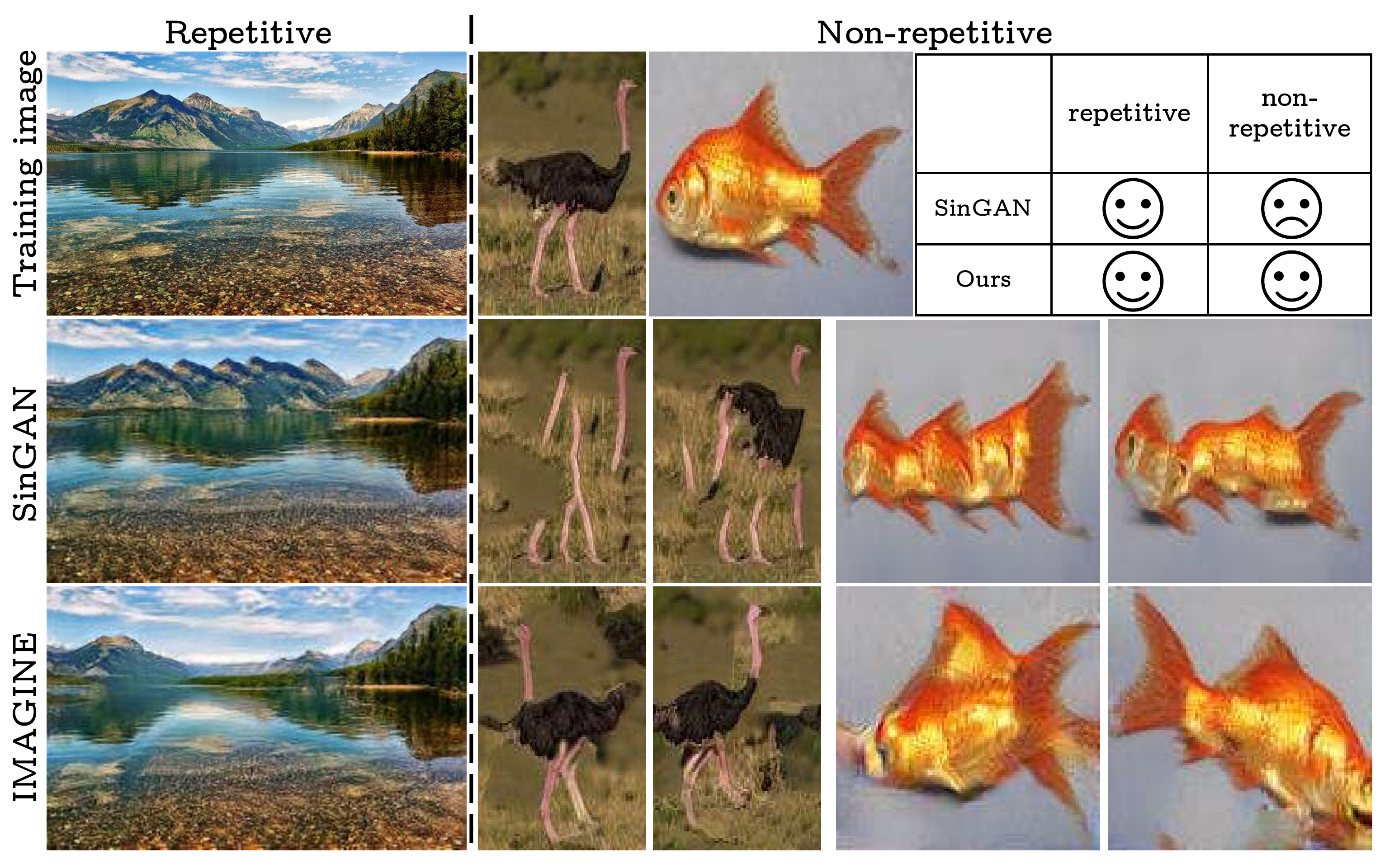}
\end{center}
   \caption{Image generation by SinGAN~\cite{rottshaham2019singan} and our IMAGINE. SinGAN only works well on repetitive images that contain repetitive structures such as mountains, but fails on non-repetitive images containing large semantic structures such as the ostrich and the goldfish. IMAGINE can generate higher quality and more diverse images on both types.}
\label{fig:teaser2}
\end{figure}

With the popularity of generative adversarial networks (GANs)~\cite{goodfellow2014generative, arjovsky2017wasserstein, karras2019style,mirza2014conditional, brock2018large},  two categories of GAN-based approaches have been proposed for editing a specific image and/or synthesizing its variations.
The first one is based on GAN projection (or inversion)~\cite{pan2020dgp,Shen_2020_CVPR,bau2019seeing,gu2020image}, which consists of projecting the target image into the latent space of the generator, jittering the projected latent code, and synthesizing a new image. 
However, there are several limitations to this type of approach. First, the projection step is far from accurate often resulting in unrealistic images due to the projected latent code falling outside of the learning distribution~\cite{Shen_2020_CVPR}. 
% \kr{I feel even if projection is not accurate it's fine as we are trying to generate variation. Maybe we need to specify from being accurate we don't mean identity but realism }. 
%The projected latent code might fall outside of the learning distribution~\cite{Shen_2020_CVPR} and could generate unrealistic images. 
Second, the projection step can only handle the type of images that the generator is trained on. 
Since existing GANs are trained on specific domains, such as faces, or datasets containing a few classes~\cite{karras2019style,Shen_2020_CVPR,odena2017conditional,zhu2020indomain,bau2020semantic,gu2020image}, the projection-based approaches cannot generalize. 
%Because most existing GANs are trained on specific domains, such as faces, or datasets containing a few classes~\cite{karras2019style,Shen_2020_CVPR,odena2017conditional,zhu2020indomain,bau2020semantic,gu2020image}, one needs different models to support severely limits the possible set of target images.
%
The second category is to learn individual GANs for individual target images, e.g., SinGAN~\cite{rottshaham2019singan}.
SinGAN leverages image patches from the target image to learn image internal statistics.
While enforcing patch consistency across resolutions enables the synthesis of multi-scale structures, SinGAN is best suited for the synthesis of repetitive images, like the mountain in Fig. \ref{fig:teaser2}.
However, as the patch-based learning is less effective in capturing high-level semantics, it struggles to synthesize images of non-repetitive objects. As shown in the second row of Fig. \ref{fig:teaser2}, properties like object identity, shape, or consistency among parts are lost in their results.

This problem can be solved by model inversion of a pre-trained deep classifier which 
can capture multi-scale abstract features. Model inversion seeks to determine the image responsible for a particular classifier prediction~\cite{mahendran2015understanding,yosinski2015understanding,nguyen2016synthesizing,yin2020dreaming}\footnote{In the remainder of the paper, we use ``model inversion'' to refer to ``classifier inversion''.}. Given the prediction, an optimization is performed to synthesize the image that best explains it. 
Although several works have shown that introducing strong regularizers~\cite{mahendran2015understanding,nguyen2016synthesizing,yin2020dreaming} can improve the quality of synthesized images, the quality is still not comparable to GANs. More importantly, all these methods suffer from {\it lack of specificity\/}. They simply generate random images with the appearance characteristics of the target class, but they cannot synthesize variations of a specific image.

In this work, to overcome the limitations of these previous methods, 
%either projection and patch-based of GAN-based learning or lack of specificity of model inversion, 
we introduce the concept of {\it IMAge-Guided model INvErsion\/} (IMAGINE). IMAGINE leverages the knowledge of image semantics from a pre-trained \emph{classifier} (e.g., ResNet-50 pre-trained on ImageNet~\cite{deng2009imagenet}) to generate semantically meaningful variations of a target.
Specifically, we feed the target image into the classifier and optimize for a new image that best explains its class prediction, under the constraint that the features of the synthesized image must match those of the target at various network layers.
% \footnote{In the remainder of the paper, we use ``model inversion'' to refer to ``classifier inversion''.}.
%
This is then complemented by the adversarial training to further improve the realism of the synthesized images. However, unlike GANs, our adversarial training is conducted by alternating the optimization of the synthesized image and that of the discriminator weights. 
% We then use adversarial training to further improve the realism of the synthesized images. Unlike GANs, our adversarial training is conducted by alternating the optimization of the synthesized image and that of the discriminator weights. 
%In order to speed up the time expense largely expended by adversarial training, we adopt a two-stage training strategy. The adversarial loss is only introduced starting from the second stage in order to minify its time. 

The resulting algorithm enables the
synthesis procedure to simultaneously 1) enforce semantic constraints during the synthesis, 2) produce realistic images, without the need for a separate generator training, and 3) allow multifaceted control over the image to be synthesized. When compared to the SinGAN, it captures much more abstract high-level image semantics, such as object shape or identity. This enables the successful synthesis of non-repetitive images, such as the objects of Fig.~\ref{fig:teaser1} and \ref{fig:teaser2}. 
Because the feature matching constraints are based on the matching of distributions, rather than minimizing geometric distances between feature maps~\cite{pan2020dgp,shocher2020semantic}, IMAGINE leads to more visual variations in the synthesized results (Fig.~\ref{fig:teaser1}). We show that, by manipulating the target feature statistics at the different levels of the network, it is possible to recreate objects with modified semantics, such as different shapes (Fig.~\ref{fig:contour}). Furthermore, through the use of attribution functions~\cite{selvaraju2017grad,sundararajan2017axiomatic}, it is even possible to manipulate the location of the target objects in the synthesized image (Fig.~\ref{fig:position}). Several other applications of IMAGINE are also discussed.

% \cl{need to define "attribution function". Not clear this way}
% \cl{Should have a paragraph summarizing our contributions.} \kr{I agree, maybe to save space the previous para can be converted as contribution}
% \pw{attached below}

Our paper makes four main contributions. (1) To the best of our knowledge, this is the first attempt to utilize model inversion for image-guided synthesis problem. (2) We introduce adversarial training to improve the image quality of optimization-based model inversion. (3) We perform extensive quantitative and qualitative evaluations to demonstrate the superiority of IMAGINE over GAN counterparts and its generalizability across different image domains including single objects in front of simple backgrounds, complex scenes, and non-stationary textures. (4) We show multiple multifaceted image control applications including a newly proposed object position control task.

\section{Related Work}
%{\bf Model inversion.} 
\noindent\textbf{Model inversion.} The goal of model inversion in vision tasks is usually to synthesize an image that maximizes the likelihood of either a network prediction or the response of some intermediate network unit. This includes the generation for adversarial attacks~\cite{fredrikson2015model,goodfellow2014explaining}, where a real image is used to initialize the optimization and the latter only seeks a prediction that is imperceptible under some constraints~\cite{szegedy2013intriguing,su2019one}. Another application would be explainable AI, where the image is initialized with noise and the goal is to determine what image structures elicit different network predictions~\cite{simonyan2013deep}. Several works in the literature have shown that the addition of different regularizers enables the synthesis of realistic images~\cite{mahendran2015understanding,nguyen2016synthesizing,nguyen2017plug}. Recently, DeepInversion~\cite{yin2020dreaming} used a regularizer based on feature matching to improve synthesis quality and showed its benefits for data-free knowledge distillation and other high-level vision tasks. These methods either produce images indistinguishable from the target image (attacks) or suffer from lack of specificity, i.e. synthesized images do not look similar to a target image. Our approach allows us to generate different variants of target image which are non-identical to target image and look real at the same time. %This is unlike the proposed image-guided model inversion.
%do not enable the synthesis of images similar to a target image.

% \cl{The following section is a bit repetitive with the intro}

\vspace{0.5em}
\noindent\textbf{Image-guided image generation.} There are several types of image-guided synthesis techniques. The most popular is GAN-inversion that uses a target image to guide the sampling of a trained GAN, attempting to sample similar images by inverting a pre-trained GAN generator~\cite{abdal2019image2stylegan,pan2020dgp,Shen_2020_CVPR,bau2019seeing,gu2020image}. They first project the target image into the latent space of the generator by minimizing certain reconstruction losses, then jitter or edit this projection so as to synthesize a new image. However, generator inversion is a difficult problem. It is hard to guarantee that the latent space projection of the target falls within the training distribution and find a semantic jittering direction.
Although some improvements have been proposed with better reconstruction quality including finetuning the generator during projection~\cite{pan2020dgp}, using multiple latent codes~\cite{gu2020image}, these literature only work on specific domains such as faces or datasets of a few classes~\cite{abdal2019image2stylegan,Shen_2020_CVPR,zhu2020indomain}, or produce inadequate results~\cite{pan2020dgp}, since capturing the distribution of highly diverse datasets with multiple object classes is still a challenging problem.
Finetuning the generator is also not space-efficient because it needs a specific GAN trained for each target image in order to work well. A more promising alternative is to train a generator specifically for image-guided synthesis. For example, \cite{shocher2020semantic} proposes to feed the target image to a pre-trained classifier and introduces the resulting features into the layers of a generator, trained to simultaneously replicate the classifier features and produce realistic images. While improving on GAN inversion, these methods still have significant limitations. Training a GAN to support the synthesis of images from many classes is still a difficult problem~\cite{brock2018large}, in particular when those are relatively fine grained and require representations with multiple semantic levels, such as ImageNet~\cite{deng2009imagenet}. 

Recently, some work~\cite{rottshaham2019singan,Shocher_2019_ICCV} propose to bypass these difficulties by leveraging internal statistics of patches within the image to train a patch-based GAN from scratch. These methods can be image conditional, i.e. take an image as input, in which case they are best suited for image processing operations, such as upsampling or retargeting~\cite{Shocher_2019_ICCV}, or unconditional, i.e. take noise as input, in which case they can perform image-guided synthesis~\cite{rottshaham2019singan}. Nevertheless, due to the patch-level learning, they have limited ability to capture high-level semantics, such as object parts and consistency among them, e.g., that `an ostrich has two legs and a neck that are connected by a torso.' As a result, they have limited ability to synthesize images that lack self-similar repetitive structure, such as the objects of Fig.~\ref{fig:teaser2}. 
Pyramid model of \cite{rottshaham2019singan} fixes the overall structure of images but loses variations.
By leveraging a deep representation with many levels of semantic abstraction not just patch-level, our image-guided model inversion offers a natural solution to this problem. 

\section{Proposed Method}
In this section, we first describe existing model inversion work for image synthesis and their limitations. After that, we introduce our image-guided model inversion work for image synthesis which overcomes the limitation of previous approaches.

\begin{figure}[t]
\setlength{\abovecaptionskip}{-2.0pt}
\setlength{\belowcaptionskip}{-0.5pt}
\setlength{\tabcolsep}{2pt}
\begin{center}
\includegraphics[width=0.6\linewidth]{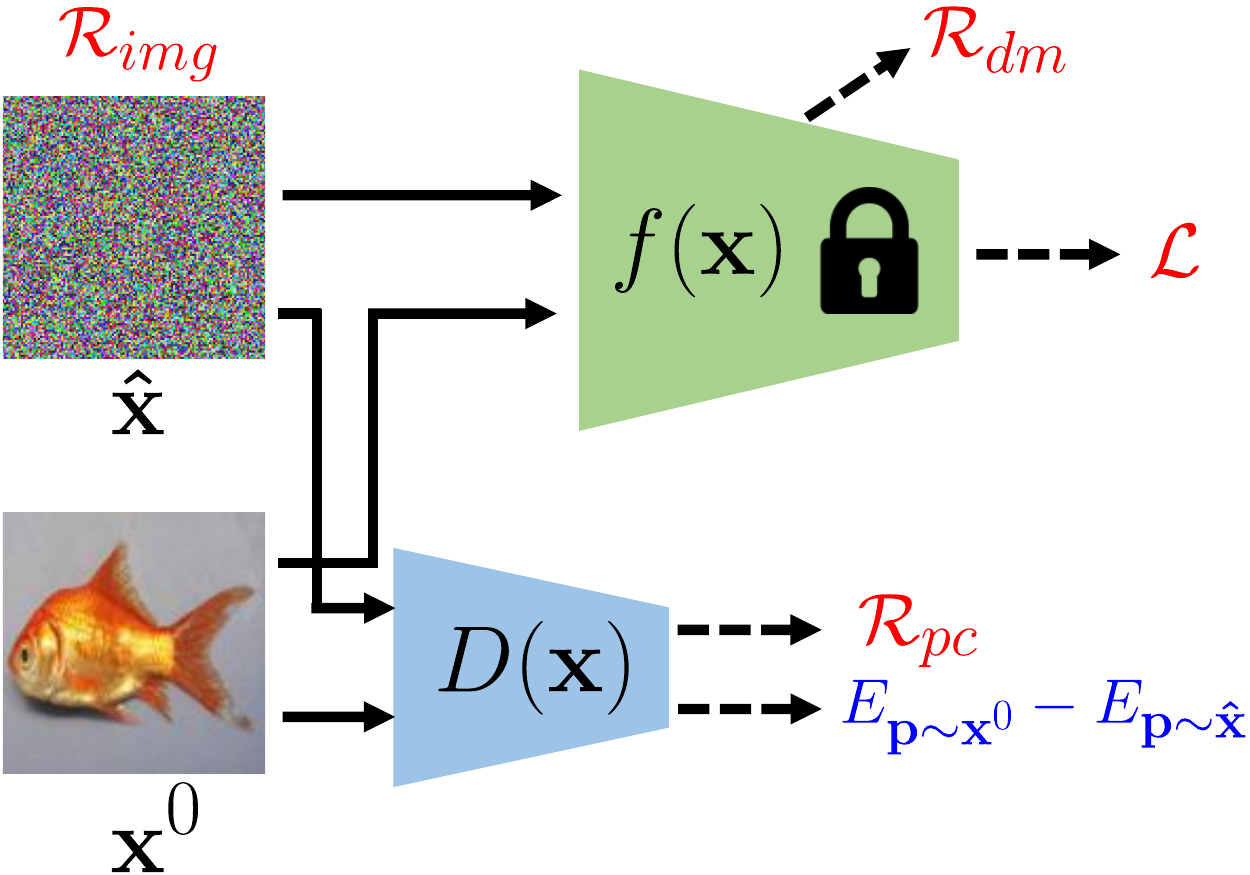}
\end{center}
   \caption{The framework of IMAGINE. The red characters represent losses for the input noise whereas blue for the discriminator. Bold arrows mean the data flow and dotted arrows loss computation.}
\label{fig:framework}
\end{figure}

\subsection{Image Synthesis by Model Inversion}

A classifier $f(\mathbf{x})$ maps images $\mathbf{x} \in \mathcal{X}$ into classes $y \in \mathcal{Y} = \{1,...,C\}$, according to $y = \argmax_{\hat{y}}  f_{\hat{y}} (\mathbf{x})$. In this work, we consider classifiers implemented with convolutional neural networks (CNNs). Model inversion aims to synthesize an image $\mathbf{x}$ that elicits a class response $y^*$ from the classifier. This is formulated as
\begin{equation}
    \mathbf{x}^* = \argmin_\mathbf{\hat{x}} {\cal L}(f(\mathbf{\hat{x}}), y^*) + {\cal R}(\mathbf{\hat{x}}),
\label{equ:opt1}    
\end{equation}
where ${\cal L}$ is the cross-entropy loss, ${\cal R}$ a regularization term, and $\mathbf{\hat{x}}$ is initialized randomly. 

Model inversion has long been used in the explainable AI literature, to visualize the patterns most representative of particular class predictions~\cite{mahendran2015understanding,simonyan2013deep,yosinski2015understanding,nguyen2016synthesizing}. These works have shown that the addition of the regularizer ${\cal R}(\mathbf{\hat{x}})$ is important to avoid highly non-realistic images. A popular choice
is to combine the total variance (TV) loss~\cite{gottlieb1998total} and the $L_2$ norm of $\mathbf{x}$ to steer $\mathbf{x}$ away from unrealistic
images with no discernible visual information~\cite{yin2020dreaming}, i.e. use
\begin{equation}
{\cal R}_{img}(\mathbf{\hat{x}}) = \alpha R_{TV}(\mathbf{\hat{x}}) + \beta ||\mathbf{\hat{x}}||^2,
\label{equ:reg} 
\end{equation}
where $\alpha$ and $\beta$ are scaling factors.

Recently,~\cite{yin2020dreaming} has shown that realism can be substantially improved by penalizing the distance between the network activations in response to $\hat{\mathbf{x}}$ and the average activations over the data used to train the network. This is denoted as a feature distribution regularizer
% \begin{eqnarray}
%     {\cal R}_{feat}(\hat{\mathbf{x}}) &=& 
%     \sum_l ||\mu_l(\hat{\mathbf{x}}) - <\mu_l>_{\cal D}||^2 \nonumber\\
%     &+&
%     \sum_l ||\sigma^2_l(\hat{\mathbf{x}}) - <\sigma^2_l>_{\cal D}||^2
%     \label{eq:rfeat}
% \end{eqnarray}
\begin{equation}
    {\scriptstyle {\cal R}_{feat}(\hat{\mathbf{x}})=\sum_l ||\mu_l(\hat{\mathbf{x}}) - <\mu_l>_{\cal D}||^2 +\sum_l ||\sigma^2_l(\hat{\mathbf{x}}) - <\sigma^2_l>_{\cal D}||^2,}
\label{eq:rfeat}
\end{equation}
where $\mu_l(\hat{\mathbf{x}})$ and $\sigma^2_l(\hat{\mathbf{x}})$ are the vectors of feature means and variances at network layer $l$, for image $\hat{\mathbf{x}}$, and $<\mu_l>_{\cal D}$ and $<\sigma^2_l>_{\cal D}$ the corresponding averages over the dataset $\cal D$ used to train the network. 

\begin{figure*}[t]
\setlength{\abovecaptionskip}{-2.0pt}
\setlength{\tabcolsep}{2pt}
\begin{center}
\includegraphics[width=0.9\linewidth]{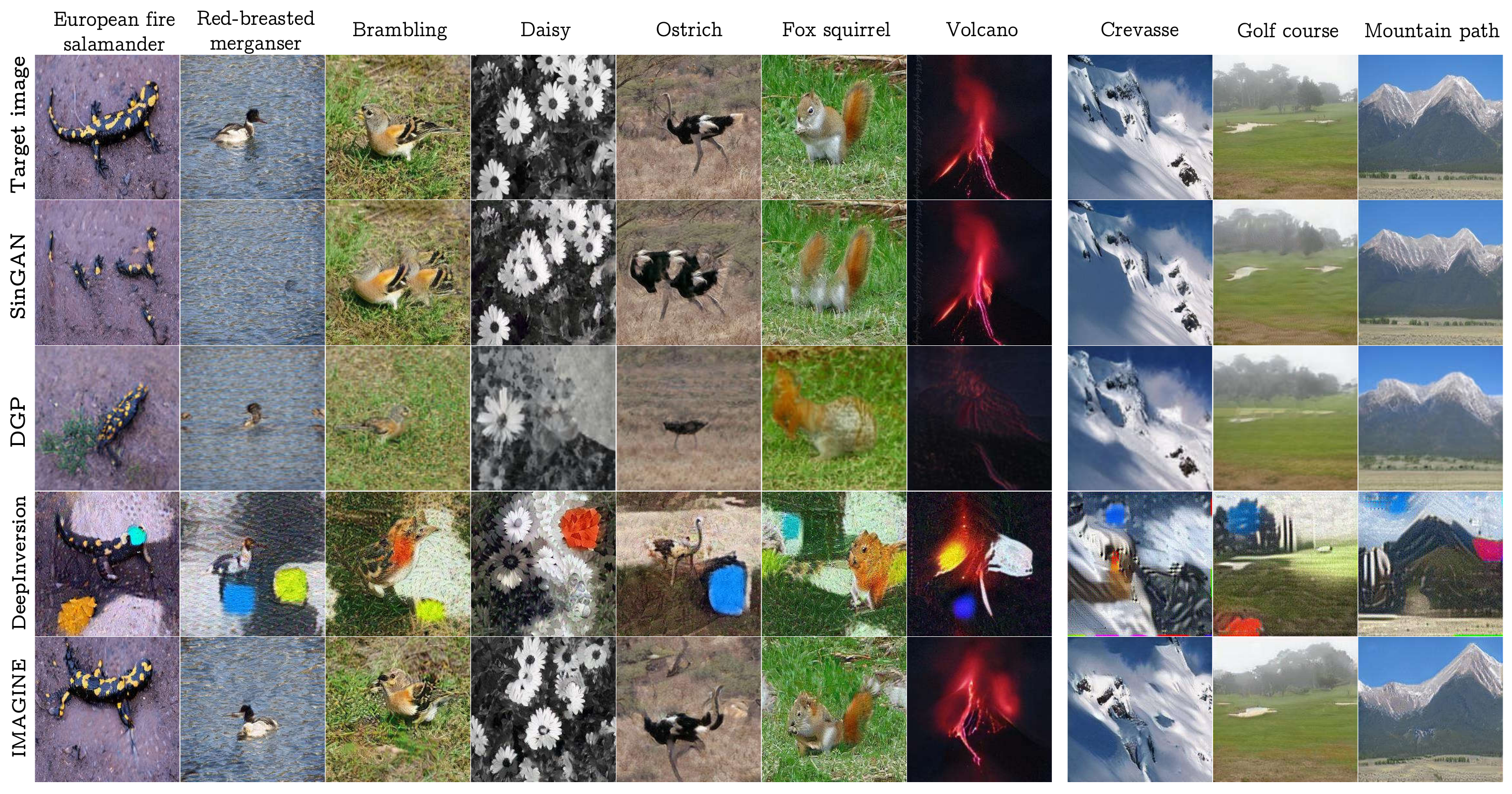}
\end{center}
   \caption{Image generation comparison of different methods (object images on left and scene images on right).}
\label{fig:object_scene}
\end{figure*}

%Although this has been shown to substantially improve the quality of synthesized images, it does not provide great control over these images. This is because $\mathbf{x}$ is matched to a set of prototypical statistics for the training set and is thus likely to be a prototypical image. It is, however, impossible to encourage the synthesis of a particular image in class $y^*$. For example, While it is possible to synthesize images of dogs, it is impossible to synthesize images of a particular dog. We denote this as {\it lack of specificity\/}. 
While improving realism, these methods suffer from {\it lack of specificity}. Although it can synthesize images of a particular class, it is impossible to synthesize image variations of a particular image. This problem has been considered in the GAN literature by inverting a pre-trained generator $G$. 
% A GAN consists of a generator $G(\mathbf{z})$ that maps a latent code $\mathbf{z}$ into an image $\mathbf{x}$. It is trained with an adversarial loss ${\cal L}_{adv}(G(\mathbf{z}), D)$, where $D$ is a discriminator between the synthesized images and a set of real images. $G$ and $D$ are trained adversarially, so that the discriminator cannot distinguish between the real and synthetic images. While GANs can synthesize images of high quality, they also suffer from lack of specificity. %Recently, some approaches have been proposed to address this problem. 
% We can mitigate this issue by inverting the generator $G$~\cite{bau2019seeing,abdal2019image2stylegan}. 
Given a target image $\mathbf{x}^0$, this consists of solving
\begin{equation}
    \mathbf{z}^* = \argmin_{\mathbf{z} \in \mathcal{R}^d} {\cal M}(\mathbf{x}^0, G(\mathbf{z})),
    \label{eq:z*}
\end{equation}
where ${\cal M}$ is a distance metric and $\mathbf{z}^*$ is the projection of $\mathbf{x}^0$ on the latent space. New images can then be generated with $\mathbf{x}^* = G(\mathbf{z}^*+\delta)$, where $\delta$ is a noise term. Although it overcomes lack of specificity, it has two problems. First, the solution of (\ref{eq:z*}) is not trivial. Simple metrics ${\cal M}$, such as the $L_2$ or perceptual loss~\cite{johnson2016perceptual}, favour similarity in terms of low-level statistics, such as color and texture, over more abstract features, such as shape or object identity. 
% This problem can be improved by, similarly to~(\ref{eq:rfeat}), matching feature at multiple semantic levels~\cite{pan2020dgp}, through minimization of the distance 
% \begin{equation}
%     {\cal R}_{disc} = \sum_l ||D^l(\mathbf{x}^0)-D^l(G(\mathbf{z}))||_1
%     \label{eq:rdisc}
% \end{equation}
% between the features produced, for the target $\mathbf{x}^0$ and
% synthesized image $G(\mathbf{z})$, by various layers $D^l$ of the the {\it discriminator\/} used to train $G$. \kr{I think we don't need to decribe the first problem in detail because anyway it can be solved by some prior work. I feel this section is becoming too long. We can directly talk about second problem and introduce single image based GAN}
A second problem is that GANs are still too difficult to train for large scale datasets involving many classes, such as ImageNet, where mode collapse and class leakage become exponentially more difficult~\cite{brock2018large}. Hence most models in the literature are trained for specific image domains, such as faces or datasets of a few classes~\cite{abdal2019image2stylegan,Shen_2020_CVPR}, and do not support the synthesis beyond these domains. \cite{pan2020dgp} proposed to solve this problem by finetuning the generator $G$ to the target image, but this fails to guarantee realistic images because the optimization process is difficult and non-trivial (see Fig. \ref{fig:object_scene}).

%.  most usually trained on a limited set of classes, 
%Ideally, if $G$ is sufficiently powerful so that the data manifold of natural images is well captured in $G$, the above objective can control the reconstructed image by dragging $\mathbf{z}$ in the latent space. However, in practice inverting a generator trained on multiple classes (like on ImageNet with 1k categories) is still challenging~\cite{pan2020dgp,brock2018large}, making it only work well on human face or some specific datasets with a few classes~\cite{Shen_2020_CVPR}. If $y^0$ is unknown, the results will also be greatly degraded. 

Single image based GANs~\cite{Shocher_2019_ICCV,rottshaham2019singan} offer an alternative to this problem. In the image-guided synthesis context, the SinGAN~\cite{rottshaham2019singan} uses the target image  $\mathbf{x}^0$ to learn a patch generator from scratch. To guarantee image consistency, both generator and discriminator operate on patches of multiple resolutions, using a pyramid structure where each stage improves on the synthesis by a stage of lower resolution. 
While the patch-based formulation bypasses the problem of training data scarcity, it has a few drawbacks. 
% First, the requirement of GAN training per image synthesis is computationally intensive. 
% For example, training a pyramid GAN of $9$ scales on $224 \times 224$ images takes $\approx2$ hours on a NVIDIA TITAN Xp. Since time and memory increase exponentially with image size, the SinGAN is impractical for many applications of single-image synthesis.
First, SinGAN needs to train a specific generator for each target image, which is not space-efficient. This is also true for GAN finetuning method~\cite{pan2020dgp}.
Second, while the multi-resolution structure constrains the organization of patches at different scales, these constraints are of low-level: the network mostly learns to tile and replicate patches at multiple scales to resemble the target image. As a result, it does not necessarily learn the high-level semantic features needed to represent object shape or long-range correlations of such features, e.g. that ``an ostrich has two legs and a neck connected by a torso" or that a ``fish has an head and a tail at opposite ends of the body." As shown in Fig.~\ref{fig:teaser2} and~\ref{fig:object_scene}, while the SinGAN is very effective for repetitive images (characterized by highly self-similar regions of different resolutions), it cannot synthesize visual concepts like object that require learning high-level semantic features capturing shape and relationships between object parts. We next introduce a new model inversion technique that aims to solve these problems.

\subsection{Image Guided Model Inversion}

To avoid the difficulties discussed above for GANs, we consider image synthesis by inversion of an image classifier. This immediately guarantees large models that encompass many classes (1k for ImageNet) and representations with multiple levels of abstraction, learned from very large datasets. However, as we saw earlier that image synthesis by inversion of an image classifier would have issues with specificity and realism. Hence, we do image guided inversion to improve these two aspects of the synthesized image. In practice, we complement $\cal R$ in~(\ref{equ:reg}) with two {\it image-guided regularizers\/}. 
%However, it has not been shown that these models can be inverted in an image guided form. 

The first is a regularizer inspired by~(\ref{eq:rfeat}).
% and~(\ref{eq:rdisc}). 
Like~(\ref{eq:rfeat}), we aim to preserve similarity between target ($\mathbf{x}^0$) and synthesized image at all levels of semantics, from class identity, to shape, contours, texture, and color. This is implemented by matching feature statistics at multiple network layers. However, to avoid lack of specificity, we seek a regularizer based on the target image alone. The goal is met by the feature distribution matching regularizer
\begin{equation}
    {\cal R}_{dm}(\mathbf{\hat{x}}; \mathbf{x}^0,\Phi) = \sum_{l\in \Phi} ||\mu_l(\mathbf{\hat{x}})-\mu_l(\mathbf{x}^0)||_2 + ||\sigma_l(\mathbf{\hat{x}})-\sigma_l(\mathbf{x}^0)||_2,
\label{equ:mm} 
\end{equation}
where $\mu_l(\mathbf{x})$ and $\sigma_l(\mathbf{x})$ are the channel-wise mean and standard deviation of feature maps at the $l$th network layer, $\Phi$ the sets of layers included in the summation.

While this regularizer enforces semantic consistency, enabling the synthesis of object-like structures, it is not sufficient to synthesize realistic images. The fine levels of image detail are usually not reproduced with enough quality to create a realistic image. To address this problem, inspired by SinGAN, we train a patch-based discriminator $D$~\cite{li2016precomputed,isola2017image} adversarially during image synthesis. Given synthesized image $\mathbf{\hat{x}}$, a patch classifier $d^*$ is trained to discriminate between  $\mathbf{\hat{x}}$ and the target $\mathbf{x}^0$, using a Wasserstein loss
\begin{equation}
    d^* = \argmax_d E_{\mathbf{p} \sim \mathbf{x}^0}[d(\mathbf{p})] -
    E_{\mathbf{p} \sim \mathbf{\hat{x}}}[d(\mathbf{p})],
    \label{eq:wass}
\end{equation}
where $\mathbf{p}$ denotes an image patch.
The image discriminator is then implemented as 
\begin{equation}
    D(\mathbf{\hat{x}}) = E_{\mathbf{p} \sim \mathbf{\hat{x}}}[d^*(\mathbf{p})].
\end{equation}
For improving training stability we use a WGAN-GP loss~\cite{gulrajani2017improved} instead of~(\ref{eq:wass}), but omit the details for brevity.
A patch consistency regularizer
\begin{equation}
    {\cal R}_{pc}(\mathbf{\hat{x}}) = - D(\mathbf{\hat{x}})
    \label{eq:rpc}
\end{equation}
is then introduced in the optimization of~(\ref{equ:opt1}).  
This encourages images that cannot be discriminated from $\mathbf{x}^0$ in terms of the visual consistency of their patches.

Finally, IMAGINE solves the optimization of~(\ref{equ:opt1}) with regularizer
\begin{equation}
    {\cal R}(\mathbf{\hat{x}};\mathbf{x}^0, \Phi) = {\cal R}_{img}(\mathbf{\hat{x}}) +
    \lambda {\cal R}_{dm}(\mathbf{\hat{x}};\mathbf{x}^0,\Phi)
    + \gamma {\cal R}_{pc}(\mathbf{\hat{x}}).
    \label{eq:Rall}
\end{equation}
where $\lambda$ and $\gamma$ are scaling factors.

\subsection{Implementation Details}

The overall framework is illustrated in Fig. \ref{fig:framework}.
In all cases, the loss of (\ref{equ:mm}) was computed on four layers: $\{conv1\_1, conv2\_3, conv3\_4, conv4\_6\}$ for ResNet-50 models, $y^*=\argmax_{\hat{y}} f_{\hat{y}}(\mathbf{x}^0)$ in (\ref{equ:opt1}), by default unless otherwise noted. The detailed structure of the discriminator $D$ is discussed in the supplementary material. Adam was used for all optimizations (learning rate $0.2$ for that of $\mathbf{\hat{x}}$ and $5e-4$ for the discriminator). Except where otherwise stated, images were synthesized with $224 \times 224$ pixels.

For all scaling weights involved in (\ref{eq:Rall}), $\alpha$ and $\beta$ of ${\cal R}_{img}$ in (\ref{equ:opt1}) are set to $1e-4$ and $1e-5$, respectively, following~\cite{yin2020dreaming}. For $\lambda$, we have found that the overall performance is stable as long as $\lambda > 1$. All results reported in this work were obtained with $\lambda = 5$. The bulk of the complexity of this optimization is in the training of the discriminator used by ${\cal R}_{pc}$. We
have found that a much more efficient solution is to use
a two-stage optimization. In the first stage, $\gamma$ is set to zero, i.e. only the feature distribution matching regularizer is used. The discriminator $D(\mathbf{x})$ is then introduced in the second stage ($\gamma = 10$), to refine the synthesized image by adversarial training. The first stage can be seen as a warm-up stage that provides a good initialization, which is refined by the second. 
Under this approach, training time is still dominated by the optimization of the patch-based adversarial loss in the second stage. However, the availability of a good initialization significantly reduces the convergence time of this optimization and, consequently, of the overall image synthesis. 
% \kr{it also improves the quality, so you should mention that, also you could mention ablation study shows the same} 
In experiments, $2000$ iterations are assigned for each stage. 
% The entire training time per training example is about $20$ minutes.
We have also tried a single stage optimization with non-zero $\lambda$ and $\gamma$ but this did not have a noticeable effect on the quality of the synthesized images and takes much more time.

\section{Experiments and Applications}
We extensively evaluate the proposed IMAGINE on different types of images against existing approaches and discuss some novel and interesting applications of our method.

\subsection{Image Synthesis}

IMAGINE is a new paradigm for image synthesis, which fills a void in the landscape of synthesis techniques. 
To the best of our knowledge, it is the first method to support image-guided synthesis by classifier inversion.
% To the best of our knowledge, it is the first method to support the synthesis of images from a large number of classes (as many as 1K for Imagenet networks), using a strong image prior (pre-trained CNN) based on feature representations with various levels of semantic abstraction, without suffering from the lack of specificity problem. \kr{we need to tone down this little bit, few other methods can do the same but not as good as us} 
Fig. \ref{fig:object_scene} compare its image synthesis performance to those of two GAN methods, DGP~\cite{pan2020dgp} and SinGAN~\cite{rottshaham2019singan}, and an image-guided version of DeepInversion~\cite{yin2020dreaming}, implemented by complementing (\ref{equ:mm}) to (\ref{eq:rfeat}) with weight factor of $5$. Just like our approach, DeepInversion method uses ResNet-50 pre-trained on ImageNet~\cite{deng2009imagenet} and Places365~\cite{zhou2017places} for object and scene images respectively as a classifier.
%These results of two model inversion methods were obtained with the ResNet-50.

The results of SinGAN, DGP, and IMAGINE are all clearly superior to those of DeepInversion. This shows the benefits of adversarial training. However, SinGAN tends to synthesize images by copying and shifting local patterns of the training image. This is effective for scenes images, e.g., ``mountain path'', containing fractal-like structures, like mountains, whose realism is not affected by this type of synthesis. However, for non-repetitive structures, such as objects like ``brambling,'' ``ostrich'' or ``fox squirrel'' (shown in Fig.~\ref{fig:object_scene}), it generates non-realistic images. In several cases, e.g., ``European fire salamander'' or ``red-breasted merganser'', the object is eliminated or ``absorbed'' into the background. DGP fails on most of the object images, perhaps because inverting a generator is always more difficult than a classifier, which causes searching a bad embedding in the latent space.

\begin{figure}[t]
\setlength{\tabcolsep}{2pt}
\begin{center}
\includegraphics[width=1.0\linewidth]{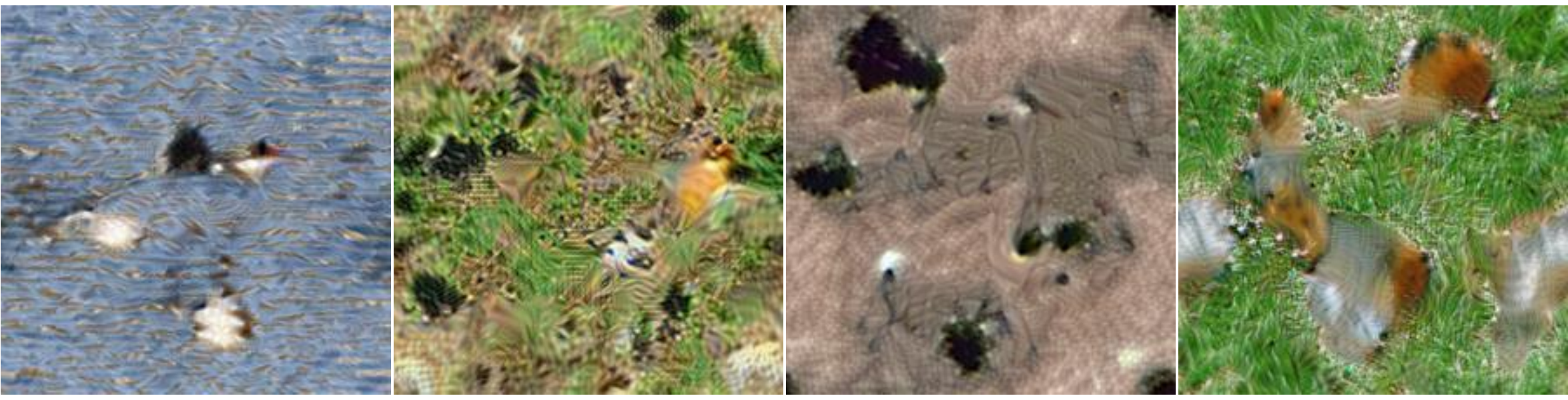}
\end{center}
   \caption{Results when we suppress the contribution of deeper layer in (\ref{equ:mm}). From left to right corresponding to ``red-breasted merganser'', ``brambling'', ``ostrich'' and ``fox squirrel'' in Fig. \ref{fig:object_scene}.}
\label{fig:ablation_higherLayer}
\end{figure}

IMAGINE is not affected by these problems, because the regularizer of~(\ref{equ:mm}) encourages the synthesis of images with activation statistics close to those of the training image {\it throughout\/} the network. This includes the activations of deeper layers, which reflect the image content in terms of higher level semantics, such as object shape and identity. This is illustrated by Fig. \ref{fig:ablation_higherLayer}, where the regularizer of (\ref{equ:mm}) was restricted only to the three earliest network layers of the four used in Fig. \ref{fig:object_scene}. Similarly to the SinGAN, the objects disintegrate and are absorbed by the background.

When the SinGAN successfully reconstructs an object or scene, e.g., ``volcano'', ``golf course'', it tends to reconstruct near replicas of the training image. This is partly due to its multi-scale structure, which enforces tight constraints on both global and local features. As a result, the network has almost no ability to synthesize diverse images whereas our approach can generate diverse images as it does not has multi-scale constraint and adversarial loss is applied to the final image. Beyond synthesizing realistic objects, IMAGINE is also able to modify the object pose (``bustard'', ``starfish''), image location (``volcano''), object contour (``island''), or even the image layout (``daisy'') as illustrated in Fig. \ref{fig:teaser1}. 

% Figure \ref{fig:texture} compares the texture synthesis performance of DANDELION with those of the SinGAN and the method of Gatys~\cite{gatys2016image}. The latter tends to break the texture pattern into pieces and fails to preserve the macro level structure of the texture (e.g. ``bubbly''). On the other hand, the SinGAN performs a relatively minor editing of the training image. Note, for example, that the four corners of the synthesized image are identical to those of the latter. This is not the case for DANDELION, which more effectively synthesizes a {\it different image of the same texture.\/} 

\begin{table}[t]
\setlength{\tabcolsep}{1pt}
\caption{Quantitative comparison of different methods on both object and scene images. Results are based on images with resolution of $224$ and presented as mean(std).}
\label{tab:IS_FID}
\centering
\scriptsize
\begin{tabular}{|l|c|c|c|c|c|c|}
\hline
& \multicolumn{3}{|c|}{Object} & \multicolumn{3}{|c|}{Scene}\\
\hline
method & IS$\uparrow$ & FID$\downarrow$ & LPIPS$\uparrow$ & IS$\uparrow$& FID$\downarrow$ & LPIPS$\uparrow$\\
\hline
% DeepInversion~\cite{yin2020dreaming} & 60.6 & - & -&- &-  &- \\
% DeepInversion$^*$ & 21.2(1.1) & 87.4(1.3) & 0.64(0.07)&7.3(0.3) & 86.4(1.7) &0.68(0.09) \\
% DeepInversion$\dagger$ & 68.5(5.0) & 65.9(2.7) & 0.52(0.10)&12.3(0.5) & 53.7(2.3) & 0.56(0.10)\\
DeepInversion~\cite{yin2020dreaming} & 68.5(5.0) & 65.9(2.7) & 0.52(0.10)&12.3(0.5) & 53.7(2.3) & 0.56(0.10)\\
SinGAN~\cite{rottshaham2019singan} & - & - & 0.24(0.11)&-  &- &0.27(0.08)\\
DGP~\cite{pan2020dgp} & 46.3(2.4) & 46.0(1.5) &0.30(0.09) &12.7(0.6)  &51.3(1.6) &0.27(0.10)\\
IMAGINE & 117.1(6.2) & 38.3(1.1) &0.46(0.09) & 21.8(0.4) &47.3(0.9) &0.43(0.13)\\
\hline
\end{tabular}
\end{table}

% \begin{table}[t]
% \caption{Comparison of user studies, mean(std) of preference percentage of IMAGINE to each listed other methods.}
% \label{tab:user_study}
% \centering
% \scriptsize
% \begin{tabular}{|l|c|c|c|c|}
% \hline
% & \multicolumn{2}{|c|}{Object} & \multicolumn{2}{|c|}{Scene}\\
% \hline
% competitor & realism & diversity & realism&diversity\\
% \hline
% DeepInversion & 100.0(0.0) & 75.0(8.2) &85.0(0.0) & 65.0(8.2)\\
% SinGAN &  60.0(7.1)& 70.0(4.1) &55.0(10.8) & 70.0(7.1)\\
% DGP &  &  & &  \\
% \hline
% \end{tabular}
% \end{table}

% \begin{table*}[t]
% \caption{Comparison of user preference ($\%$), mean$_{\text{std, confidence interval}}$ (conf. interval at 95$\%$ conf. level).}
% \label{tab:user_study}
% \centering
% \small
% \begin{tabular}{|l|c|c|c|c|}
% \hline
% & \multicolumn{2}{|c|}{Object} & \multicolumn{2}{|c|}{Scene}\\
% \hline
% competitors & realism & diversity & realism&diversity\\
% \hline
% DeepInversion/Ours & 1.0/{\bf 99.0}$_{2.0, 1.4}$ & 26.5/{\bf 73.5}$_{17.5, 6.7}$ &2.0/{\bf 98.0}$_{2.5, 2.0}$ &18.0/{\bf 82.0}$_{8.7, 5.3}$\\
% SinGAN/Ours &  29.5/{\bf 70.5}$_{10.3, 6.0}$& 17.5/{\bf 82.5}$_{6.8, 5.3}$&35.5/{\bf 64.5}$_{10.3, 6.0}$ & 19.5/{\bf 80.5}$_{10.1, 6.0}$\\
% DGP/Ours & 26.0/{\bf 74.0}$_{8.3, 6.1}$ & 28.5/{\bf 71.5}$_{7.1, 6.3}$ & 34.0/{\bf 66.0}$_{7.0, 6.6}$& 23.5/{\bf 76.5}$_{10.5, 5.9}$\\
% \hline
% \end{tabular}
% \end{table*}

\begin{table}[t]
\setlength{\tabcolsep}{1pt}
\caption{Comparison of user preference ($\%$), mean$_{\text{std, confidence interval}}$ (conf. interval at 95$\%$ conf. level).}
\label{tab:user_study}
\centering
\footnotesize
\begin{tabular}{|l|c|c|c|}
\hline
\multicolumn{4}{|c|}{Object}\\
\hline
 & DeepInversion/Ours & SinGAN/Ours & DGP/Ours\\
\hline
realism & $1.0/{\bf 99.0}_{2.0, 1.4}$ & $29.5/{\bf 70.5}_{10.3, 6.0}$ &$26.0/{\bf 74.0}_{8.3, 6.1}$\\
diversity &  $26.5/{\bf 73.5}_{17.5, 6.7}$& $17.5/{\bf 82.5}_{6.8, 5.3}$&$28.5/{\bf 71.5}_{7.1, 6.3}$\\
\hline
\multicolumn{4}{|c|}{Scene}\\
\hline
 & DeepInversion/Ours & SinGAN/Ours & DGP/Ours\\
\hline
realism & $2.0/{\bf 98.0}_{2.5, 2.0}$ & $35.5/{\bf 64.5}_{10.3, 6.0}$ &$34.0/{\bf 66.0}_{7.0, 6.6}$\\
diversity &  $18.0/{\bf 82.0}_{8.7, 5.3}$& $19.5/{\bf 80.5}_{10.1, 6.0}$&$23.5/{\bf 76.5}_{10.5, 5.9}$\\
\hline
\end{tabular}
\end{table}

The quantitative evaluation is conducted on three different datasets,
ImageNet~\cite{deng2009imagenet}, Places365~\cite{zhou2017places} and DTD~\cite{cimpoi14describing}. 
% {\bf Due to space limitation, the results on texture images, ablation studies and all experimental details are migrated to the supplementary material.} 
Tab. \ref{tab:IS_FID} presents the comparison across different methods. IMAGINE substantially outperforms all other methods, in terms of both IS~\cite{salimans2016improved} and FID~\cite{heusel2017gans}.
LPIPS~\cite{zhang2018perceptual} is calculated between all the target and synthesized image pairs to measure the patch-wised distance between them.
Higher value means synthesized image is more different and diverse from target image, and vice versa. DeepInversion obtained the best scores of LPIPS but this is meaningless as the differences between target and synthesized image pairs are due to artifacts that compromise image realism. This is evident from the low values of IS and high FID values. On the other hand, the LPIPS metric confirms that IMAGINE produces a much more diverse set of synthesized images than the SinGAN and DGP. 
% \kr{Should we talk SINGAN having lowest LPIPS score? It will again demonstrate their lack of ability to generate diverse images?} 
%This is not surprising, since the latter is meant to measure the diversity of the results, by computing the perceptual distance between training and synthesized images.
As IS and FID have to be computed based on a large number of samples and classes, it is not affordable for SinGAN due to its large computational time. Meanwhile, for target domains that only have one example available, FID is not the best metric for measuring the generation quality. Therefore, more convincing user studies are conducted to evaluate how realistic and diverse our generated results are compared with different alternatives via Amazon Mechanical Turk.

In user studies, turkers were asked to select the preferred image from a pair synthesized by two algorithms, IMAGINE vs. a competitor listed in Tab. \ref{tab:user_study}. 
% Each turker was assigned $20$ forced choice tasks and $50$ turkers were recruited for each combination of datasets and algorithm comparisons.\kr{it's not clear if each turker is given same 20 pairs or different 20 pairs} 
For each comparison, $50$ turkers were recruited and each of them was assigned $20$ same forced-choice tasks.
Two answers have to be given for each comparison with regard to two criteria: realism (select the more realistic image); diversity (select the image more different from the given target image). 
% The user interface adopted for these experiments is shown in the Supp. 
The study results are shown in Tab. \ref{tab:user_study}. A few conclusions can be drawn from the table. First, the human preferences are consistent with our qualitative observations and the quantitative results of Tab.~\ref{tab:IS_FID}. For all comparisons, IMAGINE is preferred to the competing method more than $50\%$ of the time.  Second, under the realism criterion, the IMAGINE has more advantage for objects compared to scenes which has more repetitive-like images. Third, unlike the LPIPS criterion, turkers found IMAGINE to outperform DeepInversion in terms of diversity. This is because humans consider real diversity, disregarding diversity due to artifacts that make images unrealistic.

\subsection{Position Control}

The combination of both a strong image prior, in the form of a deep classifier pre-trained over a large set of semantics, and a training image for image synthesis also offers new possibilities for image synthesis. One example is to leverage the  attribution algorithms~\cite{sundararajan2017axiomatic,selvaraju2017grad,ancona2017towards}, which produce a saliency map highlighting the image regions responsible for a class prediction. When the classifier is used for image synthesis, the attribution map can influence the synthesis itself. In particular, (\ref{equ:opt1}) can encourage images whose attribution maps meet a specification for a object location. This entails introducing in (\ref{eq:Rall}) an additional regularizer 
\begin{equation}
 {\cal R}_{loc}(\mathbf{\hat{x}};\mathbf{a}^0, y^*) = ||\mathbf{m}(f(\mathbf{\hat{x}}),y^*)-\mathbf{a}^0||_2,
 \label{eq:Rpos}
\end{equation}
that encourages a user-specified saliency map $\mathbf{a}^0$ (in our experiments a Gaussian blob at a certain image location), where $\mathbf{m}(f(\mathbf{x}),y)$ is the attribution map for the prediction of the object class $y$ in image $x$. We refer to (\ref{eq:Rpos}) as the location regularizer. In this way, rather than inferring the focus of attention of an existing image~\cite{selvaraju2017grad,zhou2016learning}, image guided model inversion enables explicit {\it control\/} of this focus of attention while generating the image.
%In this way, rather than inferring the focus of attention of an existing image, as is usually done in the saliency literature~\cite{selvaraju2017grad,zhou2016learning}, image guided model inversion enables explicit {\it control\/} of this focus of attention. 
%significant recent work in

\begin{figure}[t]
\setlength{\abovecaptionskip}{-2.0pt}
\setlength{\tabcolsep}{2pt}
\begin{center}
\includegraphics[width=1.0\linewidth]{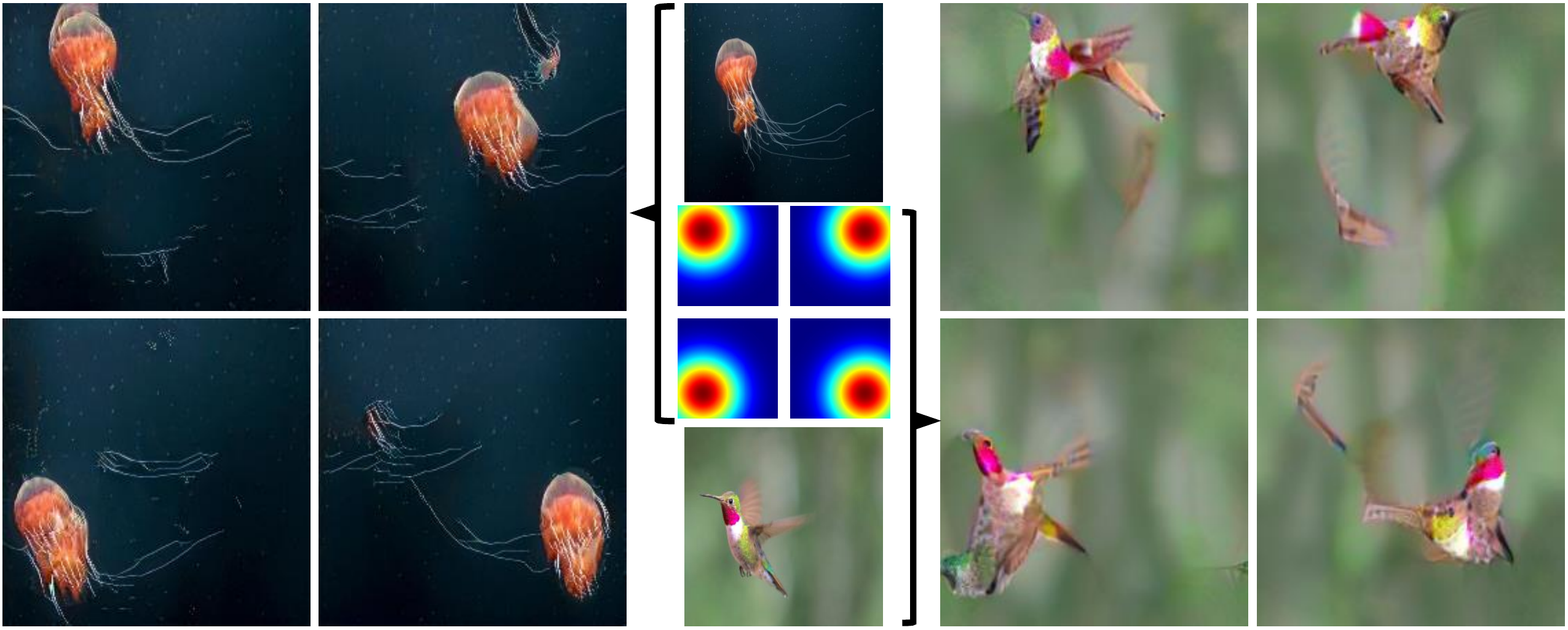}
\end{center}
   \caption{Object position control. Four different target positions (center) are used to supervise the position of objects, jellyfish (top-center) or hummingbird (bottom-center) on the generated images.}
\label{fig:position}
\end{figure}

%With the addition of this term, the loss of 
%(\ref{equ:opt3}) becomes
%\begin{equation}
%    {\scriptstyle \min_\mathbf{\hat{x}} L(f(\mathbf{\hat{x}}), y^*) + R(\mathbf{\hat{x}}) + \lambda H(\mathbf{\hat{x}}, \mathbf{x}^0; f(\mathbf{x})) + \gamma P(\mathbf{x},\mathbf{a}^0, y^*; f(\mathbf{x})) - D(\mathbf{\hat{x}}) }.
%\end{equation}
% \kr{I like the term Class Activation Maps (CAMs) rather than attribution maps especially for~\cite{selvaraju2017grad,zhou2016learning}} 
Many algorithms~\cite{selvaraju2017grad,sundararajan2017axiomatic,zhou2016learning} can be used to compute the attribution map $\mathbf{m}(f(\mathbf{x}),y)$. Since most of these boil down to back-propagating $f_y(\mathbf{x})$ to some intermediate network layer and combining with the layer activations, the attribution map can be easily computed as a side product of the optimization that is already performed to synthesize the image. This leads to an essentially cost-free mechanism for controlling object position in the synthesized image. In our experiments we have used the popular Grad-CAM~\cite{selvaraju2017grad} procedure and a weight $\nu = 10$ for (\ref{eq:Rpos}) while adding it in (\ref{eq:Rall}).
Fig.~\ref{fig:position} shows four images synthesized, for each of two training images (``jellyfish'' shown in top-center and ``hummingbird'' in bottom-center of the Figure), with four different target attribution maps $\mathbf{a}^0$ (shown in the center of the Figure). Note how, by specifying the blob $\mathbf{a}^0$, the user can control the position of either ``jellyfish'' or ``hummingbird'' object in the synthesized image. The process is effective even when the object is moved to the corner opposite to where it appears in the training image. 
%of its location The generated images are presented at the two side of figure \ref{fig:position}. Although the controlling is not too precise because the highlighting of the attribution map does not always align to the center of the object, the position can be controlled and the quality of the generated image is not sacrificed.

 %Here we try an interesting one to control the position of the object in the generated image. This is achieved by attribution map trick. Attribution map is to highlight the image regions that largely contribute to a certain prediction of the classifier. It can be produced for free from the classifier. Current popular algorithms generate the map based on the gradient of a prediction to a certain intermediate layer. If the chosen layer is deeper, the saliency map usually covers the high-level object information that provides a rough position, like the cat or dog positions in figure \ref{fig:teaser3} Grad-CAM. Inspired by this, we add another term in our objective function to encourage the position to move to the target.
 
\begin{figure}[t]
\setlength{\abovecaptionskip}{-2.0pt}
\setlength{\tabcolsep}{2pt}
\begin{center}
\includegraphics[width=1.0\linewidth]{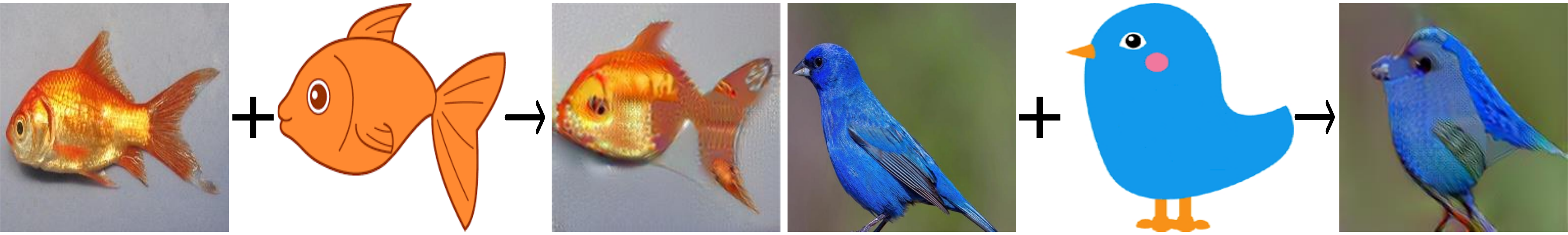}
\end{center}
\caption{Shape control: Our approach can transfer shape of a clipart to the target image.}
\label{fig:contour}
\end{figure}
 
\subsection{Shape Control}

Beyond position control, IMAGINE also enables manipulation of the semantics of the synthesized images as it leverages an image prior trained to discriminate semantics at various levels by using image representation at various levels of abstraction. Since the high and low level semantics of $\mathbf{\hat{x}}$ are controlled by the distribution matching of~(\ref{equ:mm}) at the higher and lower network layers respectively, it is possible to use different target images for low and high level semantics.
%Beyond position control, IMAGINE also enables manipulation of the semantics of the synthesized images. This is because it leverages an image prior trained to discriminate high-level semantics, like object classes, and composed of various layers  of image representation at various levels of abstraction.

Fig.~\ref{fig:contour} illustrates this idea, with an application that complements the target image $\mathbf{x}^0$ with a synthetic image, e.g. a hand-drawn or clipart image $\mathbf{x}^c$. The latter specifies a desired semantic property for the synthetic images, namely an object shape different from that depicted in the target $\mathbf{x}^0$. This exploits the fact that clipart images tend to have less texture but strong silhouette information, and are thus a good representation for shape. Given a clipart image $\mathbf{x}^c$, the feature distribution matching regularizer of~(\ref{equ:mm}) in (\ref{eq:Rall}) is modified to
\begin{equation}
 {\cal R}_{dm}(\mathbf{\hat{x}}; \mathbf{x}^c, \Phi^c) + {\cal R}_{dm}(\mathbf{\hat{x}}; \mathbf{x}^0, \Phi^r).
\end{equation}
% where $\Phi^c$ and $\Phi^r$ are the sets of layers included in the summation of~(\ref{equ:mm}). 
By including deeper and shallower layers in $\Phi^c$ and $\Phi^r$ respectively, it is possible to guide IMAGINE to synthesize images that combine the low-level semantics of the target and the high-level semantics (shape) of the clipart image. In all these examples, we use a ResNet-50 model and set $\Phi^c = \{conv4\_6\}$ and $\Phi^r = \{conv1\_1, conv2\_3, conv3\_4\}$. 
%This is to encourage the results to have a high-level similarities with the clipart where the contour is encoded, but the same texture style as the real training sample.

\subsection{Style Control}

\begin{figure}[t]
\setlength{\abovecaptionskip}{-2.0pt}
\setlength{\tabcolsep}{2pt}
\begin{center}
\includegraphics[width=0.6\linewidth]{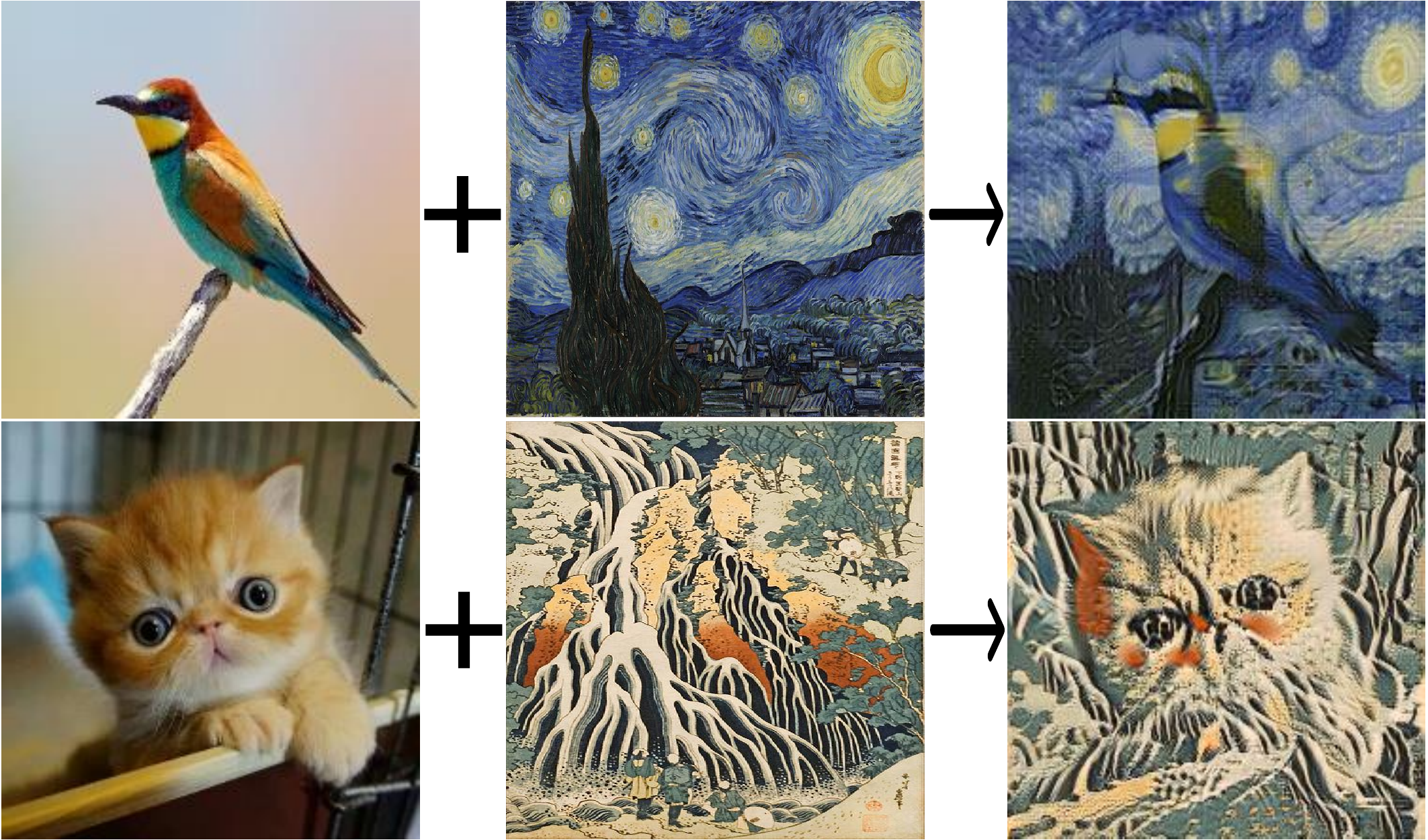}
\end{center}
\caption{Style control: our approach can transfer style of an image to target image.}
\label{fig:style_control}
\end{figure}

Inspired by the fact that style patterns usually can be encoded by the low-level layers of a classifier, similar to semantic control, IMAGINE also enables the style control by leveraging a style image to manipulate the image style of a target image~\cite{gatys2016image,johnson2016perceptual,wang2021rethinking}. Given a style image $\mathbf{x}^0$, this is achieved by replacing the distribution matching regularizer of~(\ref{equ:mm}) in (\ref{eq:Rall}) with 
\begin{equation}
 {\cal R}_{dm}(\mathbf{\hat{x}}; \mathbf{x}^0, \Phi^s),
\end{equation}
where $\Phi^s=\{conv1\_1, conv2\_3, conv3\_4\}$. 
It should be noted that here the discriminator in (\ref{eq:wass}) is fed a style image $\mathbf{x}^0$ not the real target.
% \kr{You still need a term for target image to control the shape in above equation, correct?} 
The optimization of $\mathbf{\hat{x}}$ {\it starts with the target image\/} to be stylized. 
The target image here is not included in the regularizer because \cite{du2020much} found that it is redundant when starting from the target.
The synthesized images are shown in Fig. \ref{fig:style_control}.

\subsection{Counterfactual visual explanations}
\begin{figure}[t]
\setlength{\abovecaptionskip}{-2.0pt}
\setlength{\tabcolsep}{2pt}
\begin{center}
\includegraphics[width=0.7\linewidth]{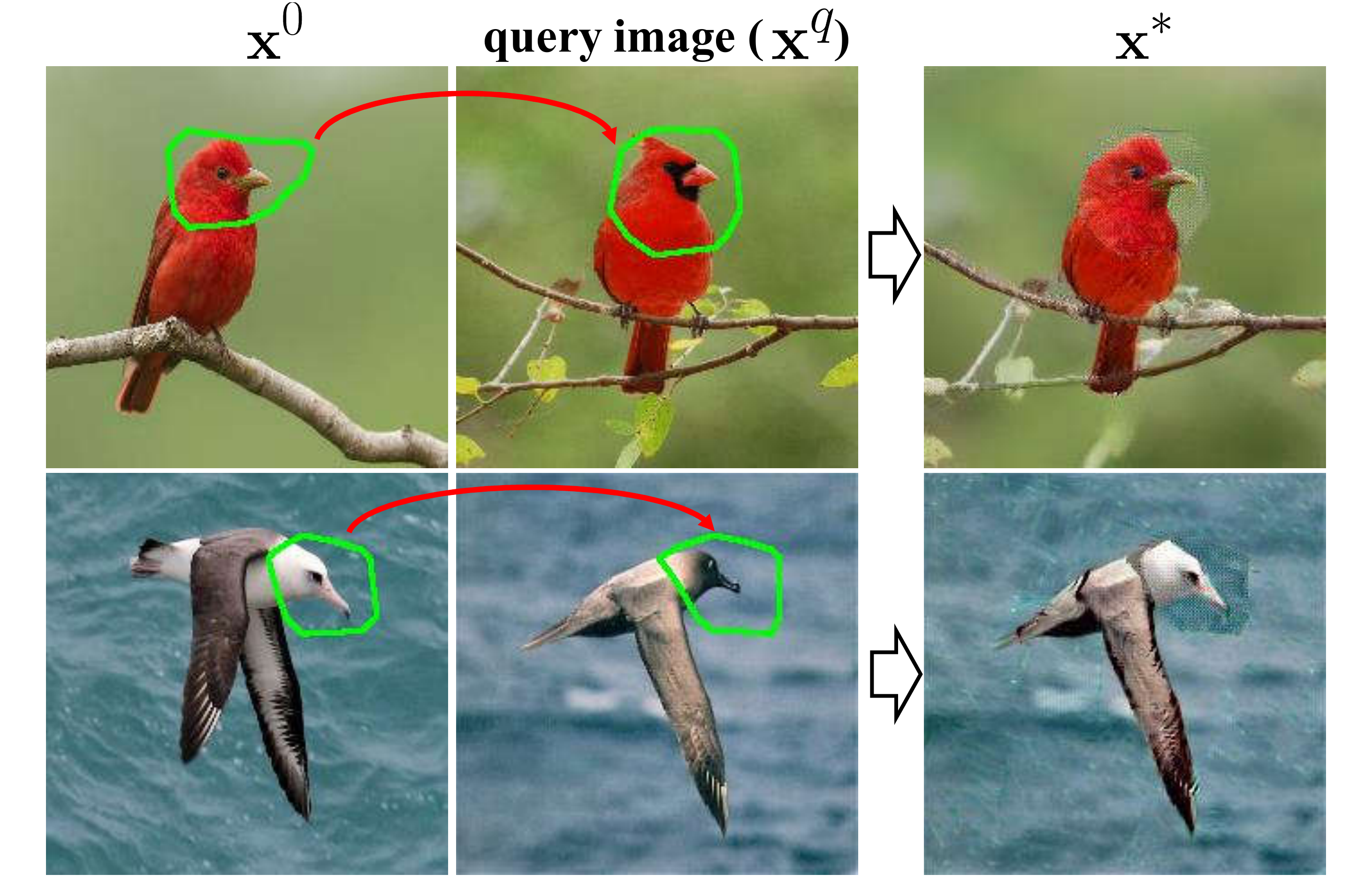}
\end{center}
\caption{Counterfactual explanations of two examples. Upper: the query image is ``Cardinal'' and counterfactual class is ``Summer Tanager''; lower: the query image is ``Sooty Albatross'' and counterfactual class is ``Laysan Albatross''.}
\label{fig:counter}
\end{figure}

For a query image ($\mathbf{x}^q$), counterfactual explanations are to answer the question with the form of ``why the prediction is class A but not class B''~\cite{goyal2019counterfactual,wang2020scout}. For example, in Fig. \ref{fig:counter} when the prediction of a classifier to the top-center image is ``cardinal'', if people ask why not a ``summer tanager'' (called counterfactual class), the system should feedback a response with the form of ``if some regions are replaced with other regions, the image would be summer tanager''. Existing literature solve this problem by 
exhaustively searching a pair of class-discriminant regions on the query image and an image $\mathbf{x}^0$ of the counterfactual class~\cite{goyal2019counterfactual}. However, it is time-consuming. \cite{wang2020scout} adopts an optimization-free method by jointly analyzing different attribution maps to localize two regions but no region replacement is introduced and changes are not done in the image space,
% \kr{it's not clear what you mean by replacement here, are you trying to say changes are not done in the image space, so it's hard to understand?}, 
which makes the explanation hard to understand. IMAGINE can solve this problem by semantically translating the counterfactual region from $\mathbf{x}^0$ to $\mathbf{x}^q$ by complementing below regularizer to (\ref{eq:Rall})
\begin{equation}
    \begin{split}
        {\cal R}_{cou}(\mathbf{\hat{x}}; \mathbf{x}^0,  \mathbf{x}^q) &= ||\mathbf{r}^q \odot (\mathbf{x}^q - \mathbf{\hat{x}})||_1 \\
    + {\cal R}_{dm}&((1-\mathbf{r}^q)\odot\mathbf{\hat{x}}; (1-\mathbf{r}^0)\odot \mathbf{x}^0, \Phi^q)
\end{split}
\end{equation}
where $\odot$ denotes element-wise multiplications and  $\Phi^q=\{conv1\_1, conv2\_3, conv3\_4\,conv4\_6\}$. $\mathbf{r}^q$ and $\mathbf{r}^0$ are the discriminant regions with the form of 0-1 mask, produced by \cite{wang2020scout}, rounded by green circles in Fig. \ref{fig:counter}, which are also side products of the classifier. The values out of the regions are $1$. Our explanation is illustrated in Fig. \ref{fig:counter} with two counterfactual explanations results. These results are based on a ResNet-50 pre-trained on CUB200~\cite{WelinderEtal2010}. So finally, taking the upper case as an example, for the center-top image, if someone asks ``why this not a summer tanager?'', our explanations would be ``if it is a summer tanager, the circled regions should look like the right image.'' 

% \paragraph{Limitation.}~Since the features extracted by the classifier can be non-robust based on the study of adversarial attack, one limitation of IMAGINE we noticed is that sometimes some ghosts are generated, e.g. ``fox squirrel'' in Figure \ref{fig:object_scene}. We visualized the attribution map of the results and found the classifier treats those ghosts as background and hence ignores them (also can be observed in Figure \ref{fig:position}). This can be potentially solved by more foreground-sensitive robust classifier.

% improving the robustness of classifier by adversarial training so that the features can be less brittle and more foreground-sensitive. 

\section{Conclusion}
In this work, we have proposed the IMAGINE for image synthesis from one single training image. 
Instead of training a GAN model, IMAGINE has synthesized images by matching various levels of semantic features of a pre-trained classifier.
%
% We have shown that our approach outperforms existing GAN-based and inversion-based methods across different image domains.
%
% For future work, we plan to evacuate more unknown knowledge inside the classifier that can benefit the image synthesis which in turn might also reward improvements on classification performance.
%Although excellent results have been obtained by IMAGINE, some artifacts can not be avoided like some checkerboards appearing at the top of synthesized images, e.g. ``golf course'' in Fig. \ref{fig:object_scene} (need to zoom in for detailed observation). These artifacts are not limited to our approach and can be found on DeepInversion and SinGAN as well. This reveals that they can be caused by the classifier or the discriminator. 
%An interesting observation is that not all results suffer from these artifacts. This indicates that artifact is dependent on the training image. We would try to address this issue as our future work.  
%

%and can benefit from future more advanced classifier.

\noindent{\bf Acknowledgement} This work was partially funded by NSF awards IIS-1924937, IIS-2041009, a gift from Amazon, a gift from Qualcomm, and NVIDIA GPU donations. We also acknowledge and thank the use of the Nautilus platform for some of the experiments discussed above.

{\small
\bibliographystyle{ieee_fullname}
\bibliography{egbib}

\begin{thebibliography}{10}\itemsep=-1pt

\bibitem{abdal2019image2stylegan}
Rameen Abdal, Yipeng Qin, and Peter Wonka.
\newblock Image2stylegan: How to embed images into the stylegan latent space?
\newblock In {\em Proceedings of the IEEE international conference on computer
  vision}, pages 4432--4441, 2019.

\bibitem{ancona2017towards}
Marco Ancona, Enea Ceolini, Cengiz {\"O}ztireli, and Markus Gross.
\newblock Towards better understanding of gradient-based attribution methods
  for deep neural networks.
\newblock {\em ICLR}, 2018.

\bibitem{arjovsky2017wasserstein}
Martin Arjovsky, Soumith Chintala, and L{\'e}on Bottou.
\newblock Wasserstein gan.
\newblock {\em arXiv preprint arXiv:1701.07875}, 2017.

\bibitem{bau2020semantic}
David Bau, Hendrik Strobelt, William Peebles, Bolei Zhou, Jun-Yan Zhu, Antonio
  Torralba, et~al.
\newblock Semantic photo manipulation with a generative image prior.
\newblock {\em arXiv preprint arXiv:2005.07727}, 2020.

\bibitem{bau2019seeing}
David Bau, Jun-Yan Zhu, Jonas Wulff, William Peebles, Hendrik Strobelt, Bolei
  Zhou, and Antonio Torralba.
\newblock Seeing what a gan cannot generate.
\newblock In {\em Proceedings of the IEEE International Conference on Computer
  Vision}, pages 4502--4511, 2019.

\bibitem{brock2018large}
Andrew Brock, Jeff Donahue, and Karen Simonyan.
\newblock Large scale {GAN} training for high fidelity natural image synthesis.
\newblock In {\em International Conference on Learning Representations}, 2019.

\bibitem{cimpoi14describing}
M. Cimpoi, S. Maji, I. Kokkinos, S. Mohamed, , and A. Vedaldi.
\newblock Describing textures in the wild.
\newblock In {\em Proceedings of the {IEEE} Conf. on Computer Vision and
  Pattern Recognition ({CVPR})}, 2014.

\bibitem{deng2009imagenet}
Jia Deng, Wei Dong, Richard Socher, Li-Jia Li, Kai Li, and Li Fei-Fei.
\newblock Imagenet: A large-scale hierarchical image database.
\newblock In {\em 2009 IEEE conference on computer vision and pattern
  recognition}, pages 248--255. Ieee, 2009.

\bibitem{du2020much}
Len Du.
\newblock How much deep learning does neural style transfer really need? an
  ablation study.
\newblock In {\em The IEEE Winter Conference on Applications of Computer
  Vision}, pages 3150--3159, 2020.

\bibitem{fredrikson2015model}
Matt Fredrikson, Somesh Jha, and Thomas Ristenpart.
\newblock Model inversion attacks that exploit confidence information and basic
  countermeasures.
\newblock In {\em Proceedings of the 22nd ACM SIGSAC Conference on Computer and
  Communications Security}, pages 1322--1333, 2015.

\bibitem{gatys2016image}
Leon~A Gatys, Alexander~S Ecker, and Matthias Bethge.
\newblock Image style transfer using convolutional neural networks.
\newblock In {\em Proceedings of the IEEE conference on computer vision and
  pattern recognition}, pages 2414--2423, 2016.

\bibitem{goodfellow2014generative}
Ian Goodfellow, Jean Pouget-Abadie, Mehdi Mirza, Bing Xu, David Warde-Farley,
  Sherjil Ozair, Aaron Courville, and Yoshua Bengio.
\newblock Generative adversarial nets.
\newblock In {\em Advances in neural information processing systems}, pages
  2672--2680, 2014.

\bibitem{goodfellow2014explaining}
Ian~J Goodfellow, Jonathon Shlens, and Christian Szegedy.
\newblock Explaining and harnessing adversarial examples.
\newblock {\em ICLR}, 2015.

\bibitem{gottlieb1998total}
Sigal Gottlieb and Chi-Wang Shu.
\newblock Total variation diminishing runge-kutta schemes.
\newblock {\em Mathematics of computation}, 67(221):73--85, 1998.

\bibitem{goyal2019counterfactual}
Yash Goyal, Ziyan Wu, Jan Ernst, Dhruv Batra, Devi Parikh, and Stefan Lee.
\newblock Counterfactual visual explanations.
\newblock {\em ICML}, 2019.

\bibitem{gu2020image}
Jinjin Gu, Yujun Shen, and Bolei Zhou.
\newblock Image processing using multi-code gan prior.
\newblock In {\em Proceedings of the IEEE/CVF Conference on Computer Vision and
  Pattern Recognition}, pages 3012--3021, 2020.

\bibitem{gulrajani2017improved}
Ishaan Gulrajani, Faruk Ahmed, Martin Arjovsky, Vincent Dumoulin, and Aaron~C
  Courville.
\newblock Improved training of wasserstein gans.
\newblock In {\em Advances in neural information processing systems}, pages
  5767--5777, 2017.

\bibitem{heusel2017gans}
Martin Heusel, Hubert Ramsauer, Thomas Unterthiner, Bernhard Nessler, and Sepp
  Hochreiter.
\newblock Gans trained by a two time-scale update rule converge to a local nash
  equilibrium.
\newblock In {\em Advances in neural information processing systems}, pages
  6626--6637, 2017.

\bibitem{isola2017image}
Phillip Isola, Jun-Yan Zhu, Tinghui Zhou, and Alexei~A Efros.
\newblock Image-to-image translation with conditional adversarial networks.
\newblock In {\em Proceedings of the IEEE conference on computer vision and
  pattern recognition}, pages 1125--1134, 2017.

\bibitem{johnson2016perceptual}
Justin Johnson, Alexandre Alahi, and Li Fei-Fei.
\newblock Perceptual losses for real-time style transfer and super-resolution.
\newblock In {\em European conference on computer vision}, pages 694--711.
  Springer, 2016.

\bibitem{karras2019style}
Tero Karras, Samuli Laine, and Timo Aila.
\newblock A style-based generator architecture for generative adversarial
  networks.
\newblock In {\em Proceedings of the IEEE conference on computer vision and
  pattern recognition}, pages 4401--4410, 2019.

\bibitem{li2016precomputed}
Chuan Li and Michael Wand.
\newblock Precomputed real-time texture synthesis with markovian generative
  adversarial networks.
\newblock In {\em European conference on computer vision}, pages 702--716.
  Springer, 2016.

\bibitem{mahendran2015understanding}
Aravindh Mahendran and Andrea Vedaldi.
\newblock Understanding deep image representations by inverting them.
\newblock In {\em Proceedings of the IEEE conference on computer vision and
  pattern recognition}, pages 5188--5196, 2015.

\bibitem{mirza2014conditional}
Mehdi Mirza and Simon Osindero.
\newblock Conditional generative adversarial nets.
\newblock {\em arXiv preprint arXiv:1411.1784}, 2014.

\bibitem{nguyen2017plug}
Anh Nguyen, Jeff Clune, Yoshua Bengio, Alexey Dosovitskiy, and Jason Yosinski.
\newblock Plug \& play generative networks: Conditional iterative generation of
  images in latent space.
\newblock In {\em Proceedings of the IEEE Conference on Computer Vision and
  Pattern Recognition}, pages 4467--4477, 2017.

\bibitem{nguyen2016synthesizing}
Anh Nguyen, Alexey Dosovitskiy, Jason Yosinski, Thomas Brox, and Jeff Clune.
\newblock Synthesizing the preferred inputs for neurons in neural networks via
  deep generator networks.
\newblock In {\em Advances in neural information processing systems}, pages
  3387--3395, 2016.

\bibitem{odena2017conditional}
Augustus Odena, Christopher Olah, and Jonathon Shlens.
\newblock Conditional image synthesis with auxiliary classifier gans.
\newblock In {\em International conference on machine learning}, pages
  2642--2651, 2017.

\bibitem{pan2020dgp}
Xingang Pan, Xiaohang Zhan, Bo Dai, Dahua Lin, Chen~Change Loy, and Ping Luo.
\newblock Exploiting deep generative prior for versatile image restoration and
  manipulation.
\newblock {\em European Conference on Computer Vision (ECCV)}, 2020.

\bibitem{rottshaham2019singan}
Tamar Rott~Shaham, Tali Dekel, and Tomer Michaeli.
\newblock Singan: Learning a generative model from a single natural image.
\newblock In {\em Computer Vision (ICCV), IEEE International Conference on},
  2019.

\bibitem{salimans2016improved}
Tim Salimans, Ian Goodfellow, Wojciech Zaremba, Vicki Cheung, Alec Radford, and
  Xi Chen.
\newblock Improved techniques for training gans.
\newblock In {\em Advances in neural information processing systems}, pages
  2234--2242, 2016.

\bibitem{selvaraju2017grad}
Ramprasaath~R Selvaraju, Michael Cogswell, Abhishek Das, Ramakrishna Vedantam,
  Devi Parikh, and Dhruv Batra.
\newblock Grad-cam: Visual explanations from deep networks via gradient-based
  localization.
\newblock In {\em Proceedings of the IEEE international conference on computer
  vision}, pages 618--626, 2017.

\bibitem{Shen_2020_CVPR}
Yujun Shen, Jinjin Gu, Xiaoou Tang, and Bolei Zhou.
\newblock Interpreting the latent space of gans for semantic face editing.
\newblock In {\em Proceedings of the IEEE/CVF Conference on Computer Vision and
  Pattern Recognition (CVPR)}, June 2020.

\bibitem{Shocher_2019_ICCV}
Assaf Shocher, Shai Bagon, Phillip Isola, and Michal Irani.
\newblock Ingan: Capturing and retargeting the "dna" of a natural image.
\newblock In {\em Proceedings of the IEEE/CVF International Conference on
  Computer Vision (ICCV)}, October 2019.

\bibitem{shocher2020semantic}
Assaf Shocher, Yossi Gandelsman, Inbar Mosseri, Michal Yarom, Michal Irani,
  William~T Freeman, and Tali Dekel.
\newblock Semantic pyramid for image generation.
\newblock In {\em Proceedings of the IEEE/CVF Conference on Computer Vision and
  Pattern Recognition}, pages 7457--7466, 2020.

\bibitem{simonyan2013deep}
Karen Simonyan, Andrea Vedaldi, and Andrew Zisserman.
\newblock Deep inside convolutional networks: Visualising image classification
  models and saliency maps.
\newblock {\em arXiv preprint arXiv:1312.6034}, 2013.

\bibitem{su2019one}
Jiawei Su, Danilo~Vasconcellos Vargas, and Kouichi Sakurai.
\newblock One pixel attack for fooling deep neural networks.
\newblock {\em IEEE Transactions on Evolutionary Computation}, 23(5):828--841,
  2019.

\bibitem{sundararajan2017axiomatic}
Mukund Sundararajan, Ankur Taly, and Qiqi Yan.
\newblock Axiomatic attribution for deep networks.
\newblock {\em ICML}, 2017.

\bibitem{szegedy2013intriguing}
Christian Szegedy, Wojciech Zaremba, Ilya Sutskever, Joan Bruna, Dumitru Erhan,
  Ian Goodfellow, and Rob Fergus.
\newblock Intriguing properties of neural networks.
\newblock {\em arXiv preprint arXiv:1312.6199}, 2013.

\bibitem{wang2021rethinking}
Pei Wang, Yijun Li, and Nuno Vasconcelos.
\newblock Rethinking and improving the robustness of image style transfer.
\newblock In {\em Proceedings of the IEEE/CVF Conference on Computer Vision and
  Pattern Recognition}, 2021.

\bibitem{wang2020scout}
Pei Wang and Nuno Vasconcelos.
\newblock Scout: Self-aware discriminant counterfactual explanations.
\newblock In {\em Proceedings of the IEEE/CVF Conference on Computer Vision and
  Pattern Recognition}, pages 8981--8990, 2020.

\bibitem{WelinderEtal2010}
P. Welinder, S. Branson, T. Mita, C. Wah, F. Schroff, S. Belongie, and P.
  Perona.
\newblock {Caltech-UCSD Birds 200}.
\newblock Technical Report CNS-TR-2010-001, California Institute of Technology,
  2010.

\bibitem{yin2020dreaming}
Hongxu Yin, Pavlo Molchanov, Jose~M Alvarez, Zhizhong Li, Arun Mallya, Derek
  Hoiem, Niraj~K Jha, and Jan Kautz.
\newblock Dreaming to distill: Data-free knowledge transfer via deepinversion.
\newblock In {\em Proceedings of the IEEE/CVF Conference on Computer Vision and
  Pattern Recognition}, pages 8715--8724, 2020.

\bibitem{yosinski2015understanding}
Jason Yosinski, Jeff Clune, Anh Nguyen, Thomas Fuchs, and Hod Lipson.
\newblock Understanding neural networks through deep visualization.
\newblock {\em arXiv preprint arXiv:1506.06579}, 2015.

\bibitem{zhang2018perceptual}
Richard Zhang, Phillip Isola, Alexei~A Efros, Eli Shechtman, and Oliver Wang.
\newblock The unreasonable effectiveness of deep features as a perceptual
  metric.
\newblock In {\em CVPR}, 2018.

\bibitem{zhou2016learning}
Bolei Zhou, Aditya Khosla, Agata Lapedriza, Aude Oliva, and Antonio Torralba.
\newblock Learning deep features for discriminative localization.
\newblock In {\em Proceedings of the IEEE conference on computer vision and
  pattern recognition}, pages 2921--2929, 2016.

\bibitem{zhou2017places}
Bolei Zhou, Agata Lapedriza, Aditya Khosla, Aude Oliva, and Antonio Torralba.
\newblock Places: A 10 million image database for scene recognition.
\newblock {\em IEEE Transactions on Pattern Analysis and Machine Intelligence},
  2017.

\bibitem{zhu2020indomain}
Jiapeng Zhu, Yujun Shen, Deli Zhao, and Bolei Zhou.
\newblock In-domain gan inversion for real image editing.
\newblock In {\em Proceedings of European Conference on Computer Vision
  (ECCV)}, 2020.

\end{thebibliography}
}

\end{document}